\documentclass[10pt,twocolumn,letterpaper]{article}

\usepackage{cvpr}              

\renewcommand{\paragraph}[1]{\vspace{.5em}\noindent\textbf{#1.}}

\usepackage{xspace}

\DeclareMathOperator*{\argmin}{arg\,min}

\usepackage{soul}
\setuldepth{foobar}

\usepackage{amsmath,amssymb,amsfonts}
\usepackage{colortbl}
\usepackage{makecell}
\usepackage{mathtools}
\usepackage{multirow}
\usepackage{pifont}
\usepackage{pgfplots}
\usepackage{tabularx}
\usepackage{lipsum}

\usepgfplotslibrary{groupplots}
\pgfplotsset{compat=1.18}
\pgfplotsset{
    every axis/.append style={
    }
}
\usetikzlibrary{patterns}
\usetikzlibrary{pgfplots.statistics}
\usepackage{pgf-pie}

\definecolor{easyblue}{RGB}{33,150,243}
\definecolor{hardorange}{RGB}{255,112,67}
\definecolor{mediumgreen}{RGB}{76,175,80}
\definecolor{lightgray}{gray}{0.85}

\newcommand{\cmark}{\ding{51}}%
\newcommand{\xmark}{\ding{55}}%
\newcolumntype{Y}{>{\centering\arraybackslash}X}

\newcommand{\R}{\mathbb{R}}

\newcommand{\Framework}{{Biomechanics-aware keypoint simulation}\xspace}
\newcommand{\framework}{{biomechanics-aware keypoint simulation}\xspace}

\newcommand{\dataset}{\textsc{SimSpine}\xspace}
\newcommand{\human}{Human3.6M\xspace}

\usepackage{xspace} 

\definecolor{cvprblue}{rgb}{0.21,0.49,0.74}
\usepackage[pagebackref,breaklinks,colorlinks,allcolors=cvprblue]{hyperref}

\def\paperID{9971} 
\def\confName{CVPR}
\def\confYear{2026}

\title{\dataset: A Biomechanics-Aware Simulation Framework for 3D Spine Motion Annotation and Benchmarking}

\author{Muhammad Saif Ullah Khan \quad Didier Stricker \\
German Research Center for Artificial Intelligence (DFKI) \\
{\tt\small \url{https://saifkhichi96.github.io/research/simspine/}}
}

\begin{document}
\maketitle
\thispagestyle{empty}
Modeling spinal motion is fundamental to understanding human biomechanics, yet remains underexplored in computer vision due to the spine’s complex multi-joint kinematics and the lack of large-scale 3D annotations. 
We present a \framework framework that augments existing human pose datasets with anatomically consistent 3D spinal keypoints derived from musculoskeletal modeling. 
Using this framework, we create the first open dataset, named \dataset, which provides sparse vertebra-level 3D spinal annotations for natural full‑body motions in indoor multi‑camera capture without external restraints. 
With 2.14 million frames, this enables data-driven learning of vertebral kinematics from subtle posture variations and bridges the gap between musculoskeletal simulation and computer vision. 
In addition, we release pretrained baselines covering fine-tuned 2D detectors, monocular 3D pose lifting models, and multi-view reconstruction pipelines, establishing a unified benchmark for biomechanically valid spine motion estimation. Specifically, our 2D spine baselines improve the state-of-the-art from 0.63 to 0.80 AUC in controlled environments, and from 0.91 to 0.93 AP for in-the-wild spine tracking.
Together, the simulation framework and \dataset dataset advance research in vision-based biomechanics, motion analysis, and digital human modeling by enabling reproducible, anatomically grounded 3D spine estimation under natural conditions.
\section{Introduction}
\label{sec:introduction}

The vertebral column, together with the pelvic girdle, forms the biomechanical core of the human skeleton—bearing axial loads, enabling locomotion, and protecting the spinal cord. Comprised of over two dozen articulating vertebrae and intervertebral joints, each with multiple degrees of freedom, the spine exhibits highly nonlinear, interdependent motion patterns~\cite{white1989clinical}. Despite decades of work across anatomy, biomechanics, neuromechanics, and rehabilitation science, precise intervertebral kinematics remain debated~\cite{kapandji1971physiology,neumann2010kinesiology,galbusera2018biomechanics}, with poor agreement on ranges of motion across \textit{in vivo} studies~\cite{rozumalski2008vivo,kozanek2009range,xia2009vivo}, cadaveric experiments~\cite{oda2002vitro,shaw2015characterization}, and computational models. Among these, \textit{in vivo} imaging (fluoroscopy, MRI, X‑ray) and motion capture are most relevant for vision-based human motion research.

\begin{figure}[t]
  \centering
   \includegraphics[clip,trim={6cm 0cm 1cm 0cm},width=0.82\linewidth]{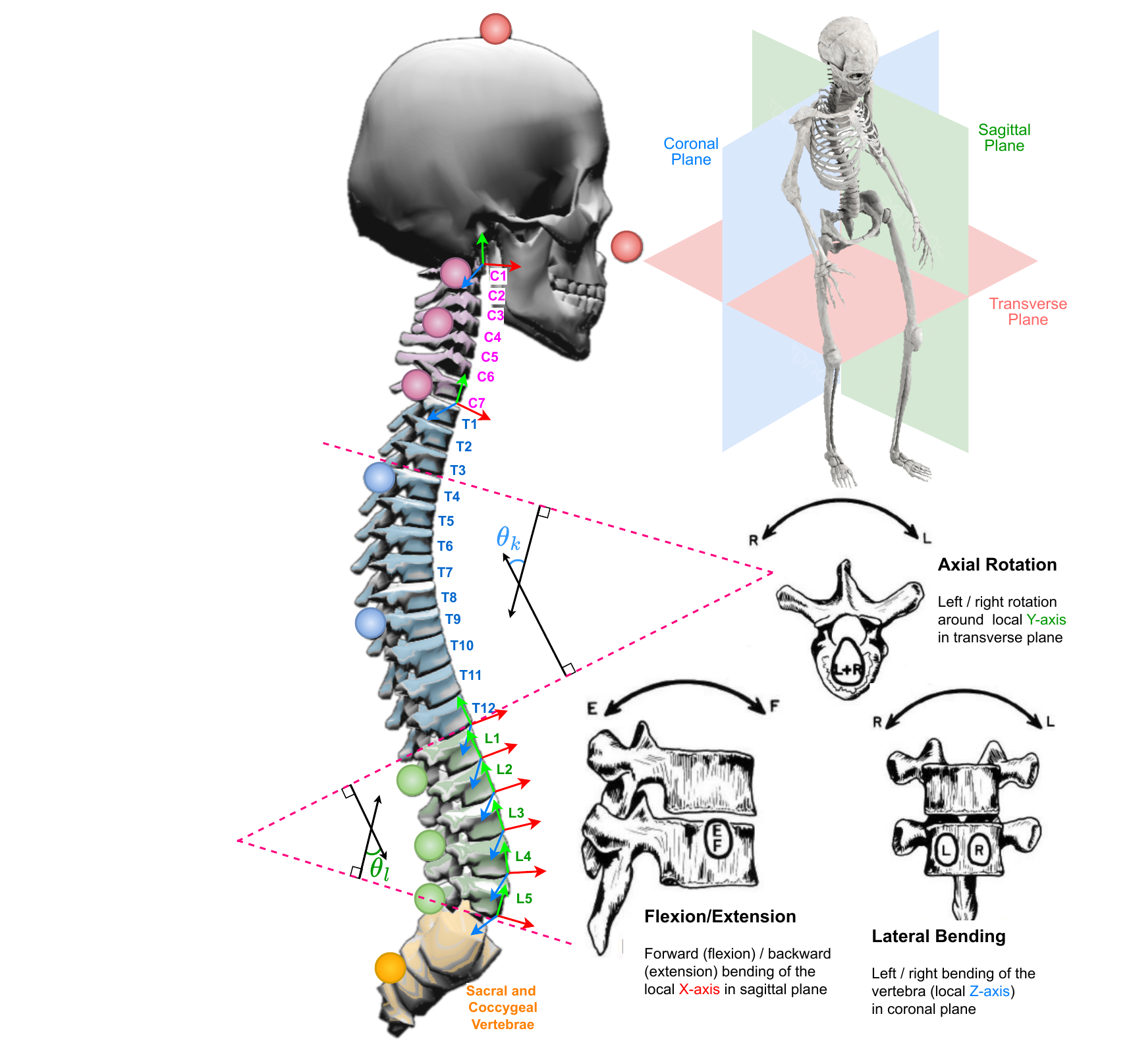}

  \caption{\textbf{\dataset annotations.} Neutral pose of the simulated spine model (left), divided into three anatomical regions—cervical (pink), thoracic (blue), and lumbar (green)—with 15 annotated landmarks: 9 along the vertebral column, 2 on the skull, 2 at the clavicle joints, and 2 on the shoulder blades \textit{(last 4 not shown)}.  Cervical and thoracic motion is limited to transition junctions (indicated by anatomical axis markers), while intermediate vertebrae remain rigidly coupled. The lumbar segment (L1–L5) is fully articulated, with intervertebral rotations simulated for all degrees of freedom (bottom right). Overall spinal curvature is characterized by thoracic kyphosis ($\theta_k$) and lumbar lordosis ($\theta_l$) angles. Motion was generated using \citeauthor{beaucage2019validation}'s musculoskeletal model~\cite{beaucage2019validation} as a function of full-body movement in \human, with 5 training and 2 validation subjects performing 15 actions.}

   \label{fig:spine-biomechanics}
\end{figure}

Traditional motion capture excels at tracking large‑scale limb motion for applications such as action recognition and human–computer interaction, but it misses subtle movements—vertebral rotations, postural sway, compensatory pelvic tilts—that influence spinal stability, load distribution, and injury risk. This limits use in sports injury prevention, ergonomics, rehabilitation, and clinical contexts.

Recent RGB approaches include \emph{marker‑assisted} tracking with perforated kinesiology tape~\cite{hachmann2023human} and a real‑world dataset/model for 2D spinal keypoint tracking~\cite{khan2025spine}. The former is difficult to deploy outside controlled settings; the latter, while scalable, remains 2D with opaque annotations and limited biomechanical verification. Thus, achieving biomechanically accurate, clinically relevant, and \emph{unconstrained} 3D estimation of healthy and pathological spine motion remains open, where “unconstrained” means not restricted to stationary subjects, close‑up unclothed views, or fixed camera angles. Following~\cite{khan2025spine}, we argue that the \textit{first step} towards solving this problem is to curate a comprehensive 3D spine motion dataset that can serve both as a baseline and a stepping stone towards finding a robust solution.

In this work, we use \textbf{\framework} to annotate an existing large-scale 3D pose dataset~\cite{ionescu2014h36m} with vertebra-level keypoints, enabling learning of spinal micro-movements from subtle whole-body posture shifts. We release: (1) \dataset, a comprehensive \textbf{dataset} containing anatomically valid 3D spinal annotations for unconstrained motions, (2) a \textbf{simulation pipeline} for generating biomechanically consistent spinal motion using musculoskeletal models, and (3) a set of \textbf{pretrained baselines} including fine-tuned 2D and 3D pose estimation models. Together, these resources provide the first open benchmark for vision-based spinal motion analysis. 

\noindent \textbf{Scope and limitations.} Annotations are simulation‑derived rather than measured \textit{in vivo}. The cervical and thoracic regions are modeled as rigid segments with motion permitted only at transition junctions (\cref{fig:spine-biomechanics}); the lumbar spine (L1–L5) is fully articulated. Motions are sourced from \human indoor activities, so appearance diversity and clinical pathologies are out of scope, and the resource is intended for research—not diagnostic—use. The biomechanically constrained 3D spine motion also supports improved realism in downstream animation and avatar synthesis tasks.

\noindent \textbf{Contributions:}
\begin{itemize}
    \item Biomechanics‑aware simulation framework that generates anatomically valid 3D spine motion data, augmenting existing human pose datasets with precise spinal labels.
    \item Pretrained spine motion baselines, including 6 fine-tuned 2D detectors for in-the-wild full-body pose estimation with state-of-the-art spine tracking, 2 monocular 3D baselines for root-relative 3D, and one multi-view 3D method for high-precision 3D in absolute metric coordinate space.
    \item Public release of the full simulation pipeline, pretrained models, and dataset—the first open 3D resource for spine motion estimation—supporting reliability, reproducibility, and benchmarking in spine‑aware pose estimation.
\end{itemize}
\section{Related Work}
\label{sec:related_work}

\paragraph{The Human Spine}
The human spine (\cref{fig:spine-biomechanics}) consists of 33 vertebrae forming an S‑shaped curve in the sagittal plane and appearing straight coronally in neutral posture~\cite{galbusera2018biomechanics}. Of these, 24 are mobile (7 cervical, 12 thoracic, 5 lumbar). Each functional motion segment permits three rotations (flexion/extension, lateral bending, axial rotation) with small, constrained translations~\cite{neumann2010kinesiology}. Reported ranges of motion (ROM) vary widely across cadaveric, in vivo, and computational studies~\cite{oda2002vitro,shaw2015characterization,galbusera2018biomechanics}, highlighting the lack of a single “ground‑truth” kinematic profile. \emph{Cervical (neck).} Most mobile region: C0–C1 mainly flexion–extension, C1–C2 primarily axial rotation, and C3–C7 distributed coupled motion~\cite{neumann2010kinesiology,lindenmann2022kinematics}. Dual‑fluoroscopy with model‑based tracking (MBT) measures sub‑degree motions but reveals task‑ and level‑specific coupling~\cite{lin2014subaxial,anderst2011validation,liu2023dynamic}. \emph{Thoracic (upper back).} Constrained by the rib cage and coronal facets, it allows moderate lateral bending and greater axial rotation in upper levels, decreasing toward the thoracolumbar junction~\cite{liebsch2018thoracic,galbusera2018biomechanics}. Upright biplanar radiography shows posture‑dependent coupling not visible in supine imaging~\cite{dubousset2005eos,humbert20093d}. \emph{Lumbar (lower back).} Large flexion/extension but limited axial rotation due to sagittal facet orientation and tall discs; rotation and lateral bending generally decrease caudally~\cite{neumann2010kinesiology,galbusera2018biomechanics}. Imaging studies report small per‑segment rotations and load‑dependent coupling~\cite{lin2014subaxial,michelini2018dynamic}.

\paragraph{Clinical Spine Tracking}
Clinical assessment of \emph{static shape and function} relies on radiographs using metrics such as Cobb’s angle and regional curvature~\cite{cobb1948outline}. Low‑dose EOS enables upright 3D reconstructions from simultaneous AP/lateral views with improved reproducibility~\cite{dubousset2005eos,melhem2016eos,humbert20093d,gajny2019quasi}. CT/MRI segmentation supports vertebra‑level analysis and planning but typically in static postures~\cite{sekuboyina2021verse,lessmann2019iterative,cheng2021automatic}. For \emph{dynamic motion}, time‑varying vertebral motion is measured with \emph{biplane fluoroscopy} plus MBT, achieving sub‑millimeter accuracy when validated against RSA~\cite{lin2014subaxial,kage2020validation,humadi2017rsa}. Dynamic MRI offers non‑ionizing alternatives for posture‑dependent deformation, but frame rate and artifacts limit kinematic fidelity~\cite{michelini2018dynamic,kulig2004assessment,liu2023dynamic}. Indirect RGB‑based gait biomarkers for scoliosis screening~\cite{kim2020meta,he2024conditional,zhou2024gait,peng2025graph} scale to large cohorts but lack vertebra‑level precision.

\renewcommand{\arraystretch}{0.9} 
\begin{table*}[ht]
    \centering
    \setlength{\tabcolsep}{3pt}
    \scriptsize
    \caption{\textbf{Comparison of existing spine datasets.} Summary of public datasets relevant to spinal imaging, scoliosis analysis, and motion estimation. Our proposed dataset uniquely provides vertebra-level 3D kinematics from RGB videos, bridging clinical imaging, biomechanics, and computer vision domains. \textit{Availability:} \cmark~Public, $\dagger$ On request, \xmark~Private.}
    \label{tab:spine-datasets}
    \begin{tabularx}{\linewidth}{p{2.1cm}cp{1.2cm}p{3.1cm}XXXc}
    \toprule
    \textbf{Dataset} &
    \textbf{Year} &
    \textbf{Input(s)} &
    \textbf{Label(s)} &
    \textbf{\# Samples} &
    \textbf{Tasks} &
    \textbf{Annotation Notes} &
    \textbf{Avail.} \\

    \midrule

    Kim et al. \cite{kim2021automatic} &
    2019 &
    Radiographs &
    Lumbar spine positions (sagittal, 2D) &
    797 images &
    Lumbar vertebrae segmentation &
    2× expert (radiologists) &
    \xmark \\

    Horng et al. \cite{horng2019cobb} &
    2019 &
    Radiographs &
    Vertebral segmentation masks (coronal, 2D) &
    595 images &
    Vertebral segmentation &
    Multi-expert (clinical) &
    $\dagger$ \\

    VinDr-SpineXR \cite{pham2021vindr} &
    2021 &
    Radiographs & 
    ROI boxes for 13 abnormalities &
    10k images / 5k cases &
    Lesion detection &
    Expert annotations &
    \cmark \\

    VerSe \cite{sekuboyina2021verse} &
    2021 &
    CT Scans &
    Vertebra centroid / segmentation masks (sagittal, 2D) &
    374 scans / 355 patients &
    Vertebral segmentation &
    Hybrid (voxel) &
    \cmark \\

    \midrule
    Scoliosis1K \cite{zhou2024gait} &
    2024 &
    Silhouette Images &
    Demographics, Scoliosis Diagnosis, 2D Body Pose &
    1.5k videos (447k frames) from 1k walking subjects &
    Scoliosis detection &
    Evaluation by medical professionals &
    \cmark \\

    USTC\&SYSU-Scoliosis~\cite{zhu2025mgscoliosis} & 2025 & 
    Radiographs, RGB & 
    Regional Cobb's angles & 
    1898 images / 1067 patients aged 10-18 &
    Scoliosis detection &
    Multi-expert agreement &
    $\dagger$ \\

    \midrule
    SpineTrack \cite{khan2025spine} &
    2025 & 
    Images (RGB) &
    9 spine keypoints (various viewpoints, 2D) & 
    33k images & 
    Keypoint Estimation &
    Hybrid, non-expert, 2D, outdoor, inconsistent &
    \cmark \\

    \rowcolor[gray]{.9} \textbf{\dataset (Ours)} &
    2025 &
    Videos (RGB) & 
    3D spine position, vertebral rotations & 
    1.56M train, 0.58M test images / 7 people, 15 actions & 
    Keypoint Estimation, Rotation Regression &
    Real+sim, biomech. plausible, indoor & 
    \cmark \\

    \bottomrule
    \end{tabularx}
\end{table*}

\paragraph{Medical Imaging-Based Spine Analysis}
Radiographs remain standard for global alignment and deformity metrics~\cite{cobb1948outline}. EOS extends these to upright 3D reconstructions~\cite{dubousset2005eos,humbert20093d}. CT provides detailed bone morphology with strong performance on segmentation benchmarks~\cite{lessmann2019iterative,cheng2021automatic,sekuboyina2021verse}. MRI captures soft‑tissue and some dynamic changes but with lower geometric fidelity for bone~\cite{michelini2018dynamic,kulig2004assessment}. Dual‑fluoroscopy with MBT yields the most accurate \emph{in vivo} kinematics, though costly and dose‑intensive; synthetic CT approaches seek to mitigate this~\cite{lin2014subaxial,kussow2025accuracy}.

\paragraph{RGB-Based Spine Motion Tracking}
RGB systems use marker-assisted tracking for structured back motion capture \cite{hachmann2023human}, but require exposed skin and controlled views. Markerless models estimate 2D spinal keypoints in natural scenes \cite{khan2025spine}, facing projection ambiguity and limited biomechanical validity. No RGB-only method yet reconstructs biomechanically consistent 3D vertebral motion under unconstrained conditions, motivating simulation-driven supervision with kinematic priors. To provide further context, \cref{tab:spine-datasets} compares key spine datasets across imaging, clinical, and vision domains. Our dataset uniquely combines real RGB sequences with anatomically constrained 3D motion.

\paragraph{OpenSim Simulation}
OpenSim \cite{delp2007opensim} is a standard platform for musculoskeletal modeling, inverse kinematics/dynamics, and forward simulation. Widely used baselines \cite{rajagopal2016full} and SimTK repositories provide spine-focused models such as: full-body and lumbar-detailed variants \cite{beaucage2019validation,christophy2012musculoskeletal}, thoracolumbar and rib-cage models with articulated T1–L5 \cite{bruno2015development}, personalizable and pediatric spines \cite{anderson2020subject,schmid2020musculoskeletal}, and cervical/neck and impact-oriented models \cite{mortensen2018inclusion,cazzola2017cervical}. Toolchains like Pose2Sim \cite{pagnon2022pose2sim} connect RGB-derived trajectories with OpenSim for simulation-ready kinematics. AddBiomechanics~\cite{werling2023addbiomechanics} provides datasets and a tool to automatically add kinematics and dynamics to joint trajectories. These provide components which our framework links with vision for large-scale image annotation.

\paragraph{3D Pose Estimation}
2D-to-3D lifting predicts absolute or root-relative 3D coordinates from monocular keypoints, most often trained on Human3.6M~\cite{ionescu2014h36m}. Foundational models range from simple MLP baselines~\cite{martinez2017simple} to temporal convolutional networks with semi-supervision~\cite{pavllo2019videopose3d}. Recent transformer-based approaches integrate motion priors for spatial–temporal consistency, such as MotionBERT~\cite{zhu2023motionbert}, MotionAGFormer~\cite{mehraban2024motionagformer}, EvoPose with structure priors~\cite{zhang2023evopose}, and PriorFormer for real-time geometric lifting~\cite{adjel2025priorformer}. While these achieve scalable full-body reconstruction, they overlook anatomical validity and intervertebral coherence.

\section{Methodology}
\label{sec:methodology}

\begin{figure*}[ht]
    \centering
    \includegraphics[width=\linewidth]{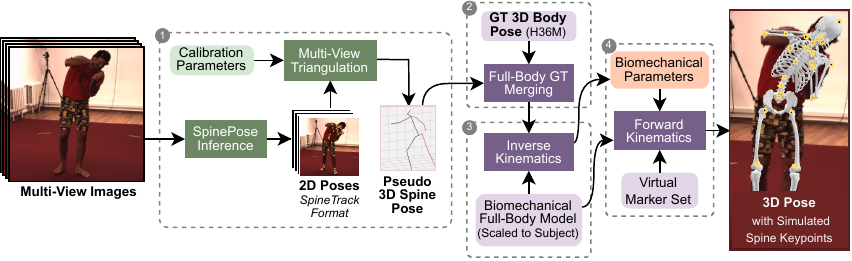}
    
    \caption{\textbf{\Framework pipeline.} From synchronized multi-view RGB, a 2D detector~\cite{khan2025spine} predicts spinal landmarks that are robustly triangulated using calibrated cameras to obtain pseudo-3D spinal keypoints. These pseudo labels are temporally aligned and merged with known \human~\cite{ionescu2014h36m} 3D markers (GT 3D Pose). OpenSim inverse kinematics (IK)~\cite{delp2007opensim} fits a subject-scaled full-body model~\cite{pagnon2022pose2sim,rajagopal2016full,beaucage2019validation} to the merged trajectories. We attach virtual markers to vertebral bodies and, using the IK joint angles and subject-specific anthropometrics, generate anatomically consistent spine keypoints via forward kinematics (FK). The pipeline also outputs biomechanical parameters (e.g., per-vertebra rotations).}
    \label{fig:spine-simulation-pipeline}
\end{figure*}

\Framework, illustrated in \cref{fig:spine-simulation-pipeline}, augments \human with sparse 3D positions of vertebral bodies, per-vertebra rotational kinematics, and subject-scaled biomechanical models. We follow the standard \human splits (train: S1, S5, S6, S7, S8; test: S9, S11), preserving \human time stamps.

\subsection{\Framework}

\paragraph{(1) Multi-view spinal detection and triangulation}
For each synchronized frame $t$ and view $v$, a pretrained detector~\cite{khan2025spine} predicts 2D spinal keypoints $\hat{\mathbf{u}}_{v,t}\in\mathbb{R}^{2\times K_s}$ (with $K_s=9$). Using calibrated intrinsics/extrinsics $\{\mathbf{K}_v,\mathbf{R}_v,\mathbf{t}_v\}$ from \human, we recover pseudo-3D points $\tilde{\mathbf{X}}_t\in\mathbb{R}^{3\times K_s}$ by robust triangulation:
\[
\tilde{\mathbf{X}}_{t}=\argmin_{\mathbf{X}}\sum_{v\in\mathcal{V}} \rho\!\left(\big\|\Pi(\mathbf{K}_v[\mathbf{R}_v|\mathbf{t}_v]\mathbf{X})-\hat{\mathbf{u}}_{v,t}\big\|_2^2\right),
\]
where $\Pi(\cdot)$ denotes perspective projection and $\rho$ is a robust penalty (Huber). We prune outliers by view-consistency and reprojection-error thresholds, and apply zero-phase low-pass filtering to suppress frame-to-frame jitter. 

\paragraph{(2) Merging with ground-truth body markers}
Let $\mathbf{Y}_t\in\mathbb{R}^{3\times K_h}$ be the \human 3D markers\footnote{Pelvis, Spine, Neck, HeadTop, and Nose from \human are not used because of overlap with our spinal points and, for the latter two points, inconsistent labels~\cite{lino2025benchmarking}.} ($K_h$ joints) in the camera/world frame. We align the pseudo-3D spinal set $\tilde{\mathbf{X}}_t$ and $\mathbf{Y}_t$ into a common OpenSim marker set $\mathcal{M}$ by (i) selecting semantic correspondences, (ii) temporal synchronization at \human frame times, and (iii) filling missing views by short-horizon interpolation. The merged markers $\mathbf{Z}_t=\{\mathbf{Y}_t,\tilde{\mathbf{X}}_t\}\in\mathbb{R}^{3\times K}$ serve as IK targets.

\paragraph{(3) Subject scaling and inverse kinematics}
We adopt a full-body OpenSim model based on Rajagopal et al.~\cite{rajagopal2016full} with lumbar spine details adapted from Beaucage-Gauvreau et al.~\cite{beaucage2019validation}; we use Pose2Sim~\cite{pagnon2022pose2sim} for data I/O and utilities. Scaling uses subject height/mass estimates derived from TRC anthropometrics (with conservative outlier trimming). IK solves, per time $t$, the weighted least-squares problem
\[
\mathbf{q}_t^\star=\argmin_{\mathbf{q}_t}\;\sum_{m\in\mathcal{M}} w_m\left\| \mathbf{z}_{m,t}-\hat{\mathbf{z}}_{m}(\mathbf{q}_t)\right\|_2^2+\lambda\|\mathbf{D}\mathbf{q}\|_2^2,
\]
where $\hat{\mathbf{z}}_{m}(\mathbf{q}_t)$ are model marker positions from FK of joint state $\mathbf{q}_t$, $w_m$ are per-marker confidences (higher for \human markers, lower for pseudo spinal points), and $\mathbf{D}$ penalizes joint-velocity/acceleration for temporal smoothness. The model includes a \emph{fully articulated lumbar spine} with three rotational DOFs at intervertebral joints from T12--L1 through L5--S1; a single 3-DOF joint at the cervicothoracic junction provides an aggregate neck DOF. Thoracic and cervical bodies beyond this aggregate are treated as rigid segments with neutral baseline curvature. This choice approximates thoracic rib-cage constraints while keeping the model identifiable with RGB-derived inputs.


\paragraph{(4) Virtual vertebral markers and forward kinematics}
We attach virtual markers to vertebral bodies (centroidal locations) and compute their 3D trajectories from the IK solution $\{\mathbf{q}_t^\star\}$ via FK. These markers define the 3D spinal keypoints distributed along the column (sacral base to lower cervical). In parallel, we export per-vertebra Euler rotations about anatomical axes (flexion/extension, lateral bending, axial rotation) as biomechanical parameters.

\begin{figure*}[t]
    \centering
    \includegraphics[width=0.64\linewidth,clip,trim={0 0 4.6cm 0}]{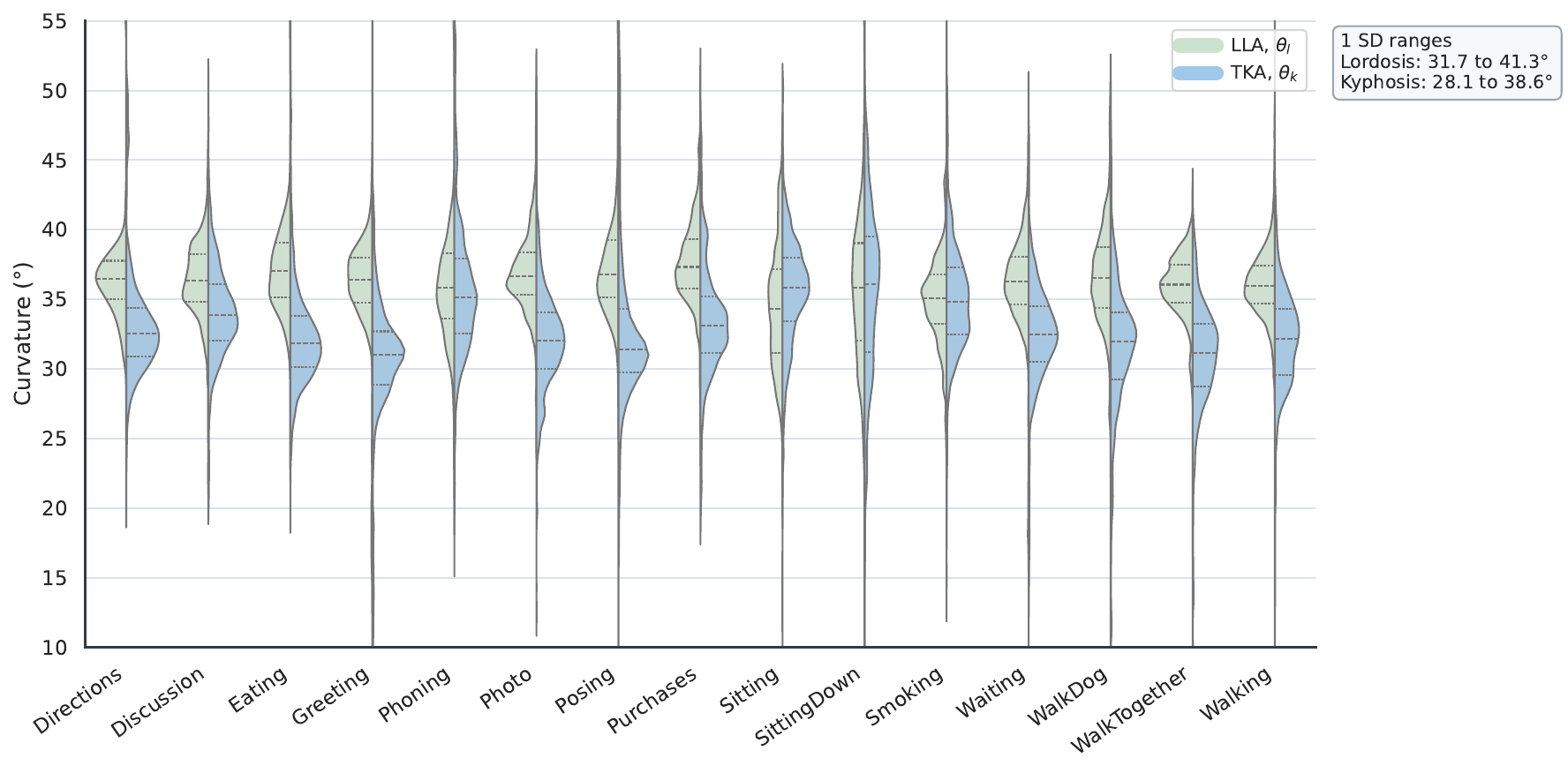}
    \begin{tikzpicture}
        \draw[dashed,gray] (0,0) -- (0,6.6);
    \end{tikzpicture}
    \includegraphics[width=0.34\linewidth]{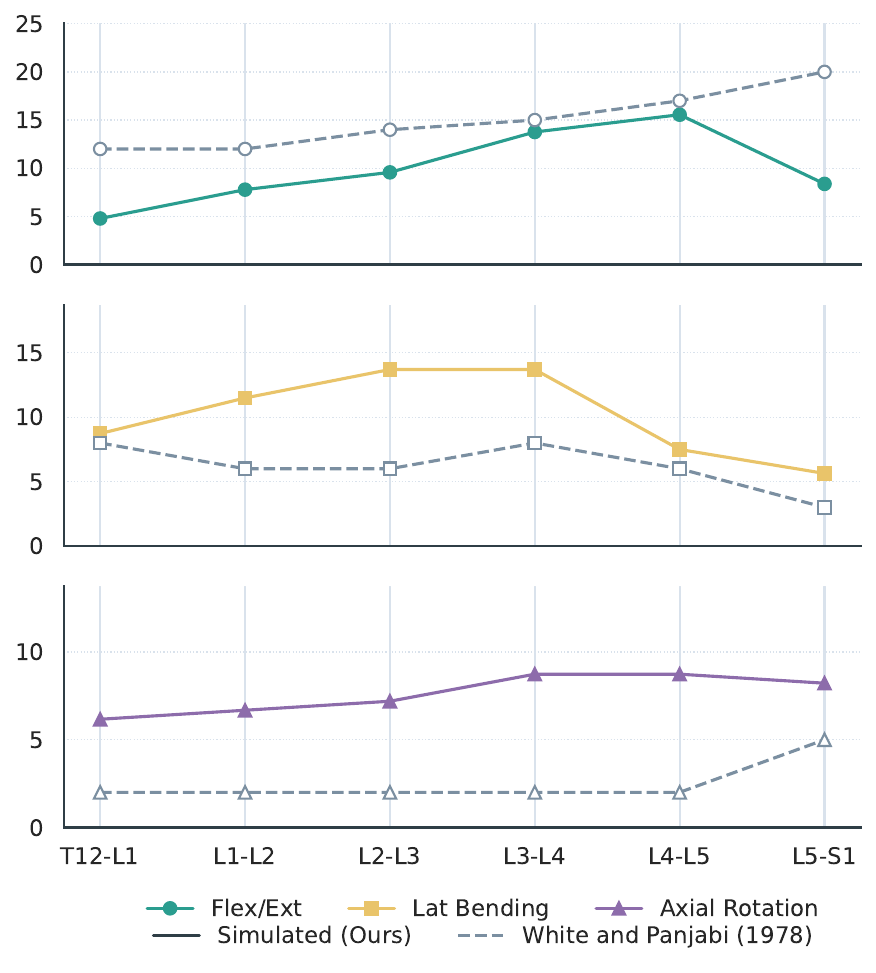}
    \caption{\textbf{Thoracolumbar spine in \dataset.} \textit{Left:} Distributions of thoracolumbar curvature across actions, defined by the Lumbar Lordotic Angle (LLA, $\theta_l$) in the lower back and Thoracic Kyphotic Angle (TKA, $\theta_k$) in the upper back. LLA and TKA average within 1~SD at 32–41° and 28–39°, respectively, indicating greater curvature in the lower spine but higher variability in the upper. Values fall within reported biomechanical ranges~\cite{lin1992lumbar,fon1980thoracic}, confirming that \dataset produces anatomically plausible curvatures and captures expected action-specific postural trends~\cite{bae2012comparison,cho2015effect,tsagkaris2022sitting,sipko2024impact,roren2024arm}. \textit{Right:} Per-vertebra range of motion (ROM) on y-axis for the three lumbar rotational DOFs. Our simulated data (solid) follows similar trends as reported by White and Panjabi (1978)~\cite{white1978basic} (dashed).}

    \label{fig:simulated-spine-analysis}
\end{figure*}
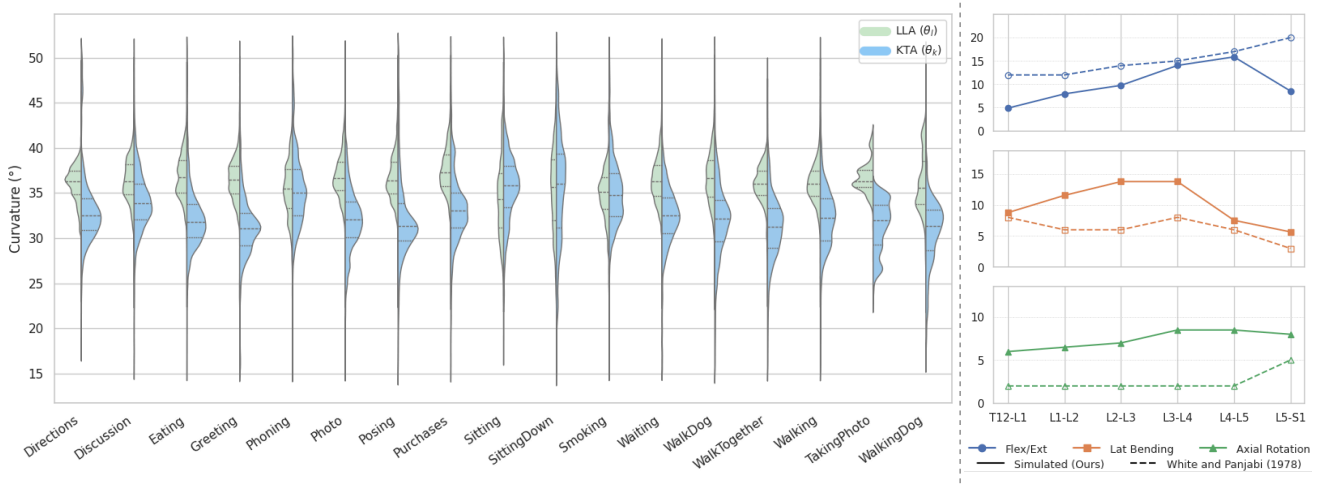

\paragraph{(5) Quality control and curation}
We (i) reject frames with implausible curvature, (ii) clamp rare angle discontinuities from gimbal wrap, and (iii) apply temporal smoothing and interpolation to fill small gaps and ensure motion continuity. Subject-specific scaled OpenSim models, marker positions, and joint angles, time synchronized with RGB frames from \human, are generated. \textbf{Markers:} 37 total, with 12 limb points from \human, 15 high-precision new points that directly model the spine, and 10 pseudo-labels on feet and face. \textbf{Kinematic axes:} 62, including 56 Euler angles. Further details are in Supplementary Sec. B.

\subsection{Biomechanical validity of simulated data}
\label{sec:validation}

We use statistical analysis to understand the simulated spine kinematics, and compare with biomechanics literature to determine the dataset's suitability for downstream tasks.

\paragraph{Thoracolumbar spine curvature} The violin plots in \cref{fig:simulated-spine-analysis} (left) show distributions of the Lumbar Lordotic Angle (LLA) and Thoracic Kyphotic Angle (TKA), computed using Cobb’s method~\cite{cobb1948outline} between L1–S1 and T3–T12 endplates, respectively. Outliers were conservatively removed. The observed distributions align with \textit{in vivo} studies and reveal activity-specific trends: seated motions (e.g., \emph{Sitting}, \emph{SittingDown}) exhibit reduced lumbar lordosis compared to standing actions (e.g., \emph{Walking}, \emph{Greeting}) due to gravitational effects on intervertebral spacing~\cite{bae2012comparison,cho2015effect,tsagkaris2022sitting,sipko2024impact}, while actions involving arm elevation (e.g., \emph{Photo}, \emph{Posing}, \emph{Greeting}) show reduced thoracic kyphosis from posterior shoulder displacement~\cite{roren2024arm}. Across all actions, mean lordosis (32–41°) and kyphosis (28–39°) fall within normative adult ranges~\cite{lin1992lumbar,fon1980thoracic}. These bounded, unimodal, and action-sensitive distributions reflect realistic postural variability in healthy adults, underscoring the biomechanical fidelity of the simulated data. Details of subject and action-specific ranges is provided in Supplementary Tab. B2.

\begin{figure}[t]
  \centering
  \includegraphics[width=\linewidth]{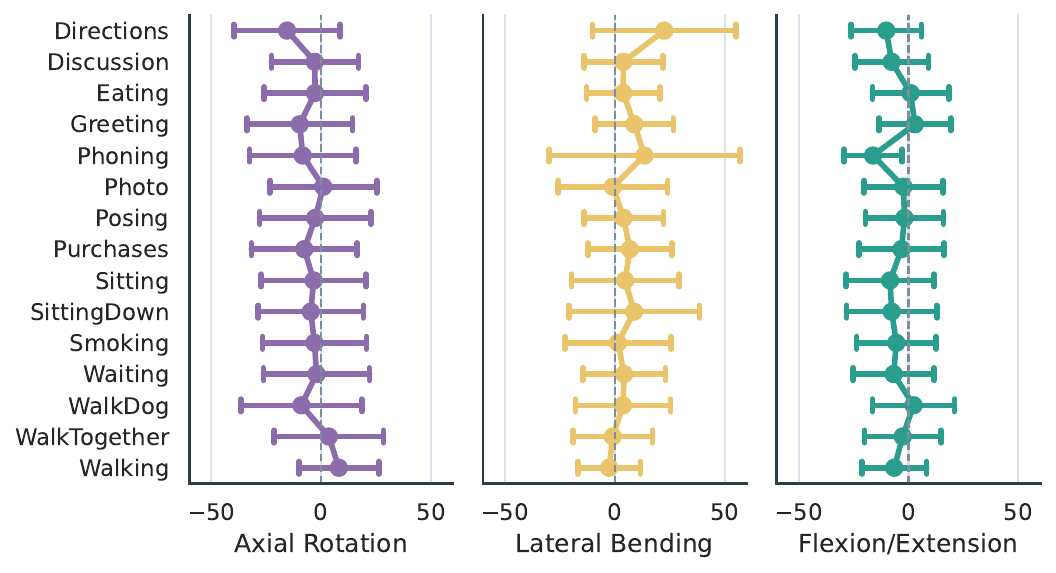}
  \caption{\textbf{Cervical spine in \dataset.} The distributions (per action) remain centered near neutral with task-dependent spread, reflecting that our model uses a single 3-DOF aggregate neck joint while keeping the thoracic/cervical bodies rigid beyond the cervicothoracic junction. This is within the neck ROM reported in \cite{doss2023comprehensive} with approximately half coverage, which indicates the presence of only small head movements in the dataset.}
  \label{fig:neck-rom}
\end{figure}

\paragraph{Lumbar spine ROM trends} 
The lumbar ROM curves in \cref{fig:simulated-spine-analysis}, (right) display known qualitative gradients: flexion/extension increases from the thoracolumbar junction toward the caudal levels with a peak near L4–L5, then reduces at L5–S1; lateral bending peaks mid-lumbar; axial rotation is modest overall and highest in mid-lumbar segments. These monotone or near-monotone trends match widely reported patterns~\cite{white1978basic} for upright motion and are difficult to reproduce with unconstrained pose-only priors; their presence suggests the IK solution respects both morphology and motion constraints. Simulated values follow similar trends as the reference data. It should be noted that exact ROM values for lumbar vertebrae are debated~\cite{savlovskis2020range}.

\paragraph{Cervical proxy}
Despite modeling the cervical spine as a single aggregate 3-DOF joint, action-conditioned neck ROM distributions in~\cref{fig:neck-rom} stay within physiologically credible envelopes and broaden for tasks with head motion. For example, the \emph{Phone} action has the largest lateral bending range indicative of a sideways head tilt typical during phone conversations. 
This supports usefulness for coarse head–neck kinematics and for supervising 2D-to-3D lifting in the absence of vertebra-resolved cervical detail.

These graphs jointly indicate that our simulation yields \emph{anatomically plausible} and \emph{action-sensitive} kinematics. This supports the dataset’s suitability as supervision for learning vertebral motion from natural movements and as a benchmark for biomechanics-aware spine motion models.

\section{Baselines and Experiments}
\label{sec:experiments}

We benchmark three tasks enabled by \dataset: (1) 2D pose estimation from RGB, (2) multiview 3D pose reconstruction, and (3) monocular 3D pose lifting. We formally define each task, describe the baselines, training methods, evaluation protocols, and present ablation studies. The aim is to provide deployment-ready, robust pretrained models, and quantitative results which future spine research can use as a reference. Known limitations are also discussed.


\paragraph{Notation}
Let $B$ be batch size, $T$ frames per clip ($T{=}1$ for image-based; $T{>}1$ for video-based), $K$ vertebral keypoints, and $C_{\text{img}}{=}3$ image channels. For each sequence, \(\mathbf{I} \in \R^{B \times T \times C_{\text{img}} \times H \times W}\) are RGB frame(s), \(\mathbf{U} \in \R^{B \times T \times K \times 2}\) are 2D keypoints, and \(\mathbf{Y} \in \R^{B \times T \times K \times 3}\) are 3D coordinates in a global camera or anatomical frame.

\subsection{2D Pose Estimation}

\paragraph{Definition}
This task aims to learn a function
\[
f_{\text{2D}}:\R^{B \times T \times C_{\text{img}} \times H \times W} \rightarrow \R^{B \times T \times K \times 2},
\]
predicting $\widehat{\mathbf{U}}$ from images with supervision $\mathbf{U}$ and visibility weights $\mathbf{V}\!\in[0,1]^{B\times T\times K}$.

\paragraph{Baselines}
We fine-tune three variants of SpinePose~\cite{khan2025spine} and compare them with representative architectures from both CNN and transformer families: HRNet-w32~\cite{sun2019deep}, RTMPose-m~\cite{jiang2023rtmpose}, and ViTPose-b~\cite{xu2022vitpose}. HRNet and ViTPose are heatmap-based methods while RTMPose is a coordinate classification method.

\paragraph{Training}
All models are initialized with pretrained weights and fine-tuned on a balanced combination of outdoor (SpineTrack) and indoor (\dataset) images, with equal samples per batch. Only 2\% of \dataset train split are used to avoid overfitting while still transferring simulation-derived knowledge. Each model is fine-tuned for 10 epochs with a three-stage curriculum where data augmentation transitions from hard to easy (see Supplementary Sec. A). Optimizers and losses follow SpinePose~\cite{khan2025spine}.

\begin{table}[ht]
  \centering
  \scriptsize
  \setlength{\tabcolsep}{2.5pt}
  \caption{\textbf{Baselines for 2D Spine Pose Estimation.} Performance of various CNN and transformer architectures on the SpineTrack and \dataset benchmarks.}
  \label{tab:baselines-spine2d}
  \begin{tabularx}{\linewidth}{lccYYYYc}
    \toprule
    & \multicolumn{2}{c}{\textbf{Dataset}}& \multicolumn{4}{c}{\textbf{SpineTrack}} & \textbf{Ours} \\
    \cmidrule(lr){2-3}
    \cmidrule(lr){4-7}
    \cmidrule(lr){8-8}
    \textbf{Method} & \textbf{Pretrain}  & \textbf{Finetune} & \textbf{AP$^{\mathrm{B}}$} & \textbf{AR$^{\mathrm{B}}$} & \textbf{AP$^{\mathrm{S}}$} & \textbf{AR$^{\mathrm{S}}$} & \textbf{AUC} \\
    
    \cmidrule{1-3}
    \cmidrule(lr){4-7}
    \cmidrule(lr){8-8}
    SpinePose-s    & \multirow{3}{*}{SpineTrack} & \multirow{3}{*}{-} & 0.792 & 0.821 & 0.896 & 0.908 & 0.611 \\
    SpinePose-m    & & & 0.840 & 0.864 & 0.914 & 0.926 & 0.633 \\
    SpinePose-l    & & & \underline{0.854} & \underline{0.877} & 0.910 & 0.922 & 0.633 \\
    \cmidrule{1-3}
    \cmidrule(lr){4-7}
    \cmidrule(lr){8-8}

    SpinePose-s-ft & \multirow{3}{*}{SpineTrack} & \multirow{3}{*}{\makecell{SpineTrack \\ + Ours}} & 0.788 & 0.815 & 0.920 & 0.929 & 0.790 \\
    SpinePose-m-ft & & & 0.821 & 0.846 & \underline{0.928} & \underline{0.937} & 0.798 \\
    SpinePose-l-ft & & & 0.840 & 0.862 & 0.917 & 0.927 & \underline{0.803} \\
    \cmidrule{1-3}
    \cmidrule(lr){4-7}
    \cmidrule(lr){8-8}
    HRNet-w32 & \multirow{3}{*}{COCO} & \multirow{3}{*}{\makecell{SpineTrack \\ + Ours}} & 0.776 & 0.806 & 0.905 & 0.918 & 0.769 \\
    RTMPose-m & & & 0.832 & 0.858 & 0.925 & 0.935 & 0.794 \\
    ViTPose-b & & & 0.835 & 0.866 & 0.921 & 0.933 & 0.794 \\
    
    \bottomrule
  \end{tabularx}
\end{table}

\paragraph{Evaluation}
On SpineTrack we report AP/AR using COCO-style OKS with body (B) and spine (S) subsets following \cite{khan2025spine}.
On \dataset we report AUC of PCK over thresholds $\tau \in [0,0.5]$, with distances normalized by the shorter side of the person bounding box. Higher is better for all metrics. Comparisons with the original SpinePose~\cite{khan2025spine} quantify the effect of consistent labeling in \dataset. 


\subsection{Multiview 3D Reconstruction}

\paragraph{Definition}
Given synchronized detections $\{\widehat{\mathbf{U}}^{(m)}\}_{m=1}^M$ and calibrated projections $\{P^{(m)}\}_{m=1}^M$, triangulate
\[
\widehat{\mathbf{Y}} \;=\; f_{\text{tri}}(\{\widehat{\mathbf{U}}^{(m)}, P^{(m)}\}_{m=1}^M),
\]
by minimizing multi-view reprojection error.

\paragraph{Baselines} 
We use a weighted linear least-squares triangulation with confidence-based outlier rejection~\cite{hartley1997triangulation,hartley2003multiple}. Two 2D detectors are compared: the zero-shot \textit{SpinePose-m} pretrained on SpineTrack~\cite{khan2025spine}, and its fine-tuned variant \textit{SpinePose-m-ft} adapted on \dataset.

\begin{table}[ht]
    \centering
    \scriptsize
    \caption{\textbf{Multiview 3D Spine Reconstruction.} MPJPE (mm) across actions and spinal regions using linear triangulation. $S_C$: Cervical, $S_T$: Thoracic, $S_L$: Lumbar, $S$: Full spine, $B$: Body, and All: Complete skeleton.}
    \begin{tabularx}{\linewidth}{lYYYYYY}
    \toprule
    \textbf{Action} &
    \textbf{${\mathrm{S_C}}$} &
    \textbf{${\mathrm{S_T}}$} &
    \textbf{${\mathrm{S_L}}$} &
    \textbf{${\mathrm{S}}$} &
    \textbf{${\mathrm{B}}$} &
    \textbf{${\mathrm{All}}$} \\
    \midrule
    Mean (GT 2D)& 12.36 & 8.80 &  5.09 &  9.00 &  7.07 &  7.85 \\
    \midrule
    Directions & 33.69 & 39.86 & 40.16 & 37.88 & 27.63 & 31.79 \\
    Discussion & 36.35 & 40.55 & 41.44 & 39.39 & 28.33 & 32.81 \\
    Eating & 38.47 & 43.04 & 35.40 & 39.48 & 27.44 & 32.32 \\
    Greeting & 30.66 & 35.19 & 33.73 & 33.29 & 25.57 & 28.70 \\
    Phoning & 42.41 & 44.64 & 39.14 & 42.43 & 27.23 & 33.40 \\
    Photo & 36.40 & 42.13 & 42.27 & 40.25 & 30.35 & 34.37 \\
    Posing & 33.48 & 37.46 & 40.14 & 36.85 & 28.02 & 31.60 \\
    Purchases & 34.81 & 35.67 & 39.63 & 36.44 & 26.64 & 30.61 \\
    Sitting & 40.94 & 43.16 & 31.21 & 39.23 & 25.42 & 31.02 \\
    SittingDown & 34.85 & 35.95 & 23.63 & 32.30 & 23.31 & 26.95 \\
    Smoking & 38.51 & 44.83 & 35.66 & 40.28 & 27.26 & 32.54 \\
    Waiting & 35.17 & 37.40 & 33.22 & 35.54 & 24.78 & 29.14 \\
    WalkDog & 35.53 & 41.44 & 40.83 & 39.31 & 29.49 & 33.47 \\
    WalkTogether & 34.05 & 43.88 & 40.40 & 39.68 & 29.28 & 33.50 \\
    Walking & 33.02 & 43.80 & 41.53 & 39.60 & 29.06 & 33.33 \\
    \midrule
    \rowcolor{gray!30}
    Mean (Finetuned) & \underline{36.50} & \underline{41.07} & \underline{37.13} & \underline{38.50} & \underline{27.27} & \underline{31.82} \\
    Mean (Zero-Shot) & 54.39 & 69.90 & 48.13 & 58.92 & 42.74 & 49.30 \\
    \bottomrule
    \end{tabularx}
    \label{tab:baselines-spine3d-multiview}
\end{table}

\paragraph{Evaluation} For triangulated poses in global world coordinates, we report Mean Per-Joint Position Error (MPJPE):
\[
\mathrm{MPJPE} \;=\; \frac{1}{B T K} \sum_{b,t,j} \,\big\lVert \mathbf{Y}^{\text{pred}}_{b,t,j,:} - \mathbf{Y}_{b,t,j,:} \big\rVert_2.
\]
Results are summarized in~\cref{tab:baselines-spine3d-multiview}. To further isolate geometric shape from global alignment,~\cref{tab:triangulation-pmpjpe} reports Procrustes-aligned MPJPE (P-MPJPE).

\begin{table}[ht]
    \centering
    \scriptsize
    \caption{\textbf{Triangulation Baseline: P-MPJPE (mm).}
    Same setup as Table~\ref{tab:baselines-spine3d-multiview}, evaluated after similarity alignment.}
    \label{tab:triangulation-pmpjpe}
    \begin{tabularx}{\linewidth}{lYYYYYY}
    \toprule
    \textbf{Action} &
    \textbf{${\mathrm{S_C}}$} &
    \textbf{${\mathrm{S_T}}$} &
    \textbf{${\mathrm{S_L}}$} &
    \textbf{${\mathrm{S}}$} &
    \textbf{${\mathrm{B}}$} &
    \textbf{${\mathrm{All}}$} \\
    \midrule
    Mean (GT 2D)     & 0.33 & 0.43 & 0.06 & 0.67 & 1.81 & 1.79 \\
    \midrule
    \rowcolor{gray!30}
    Mean (Finetuned) & \underline{19.68} & \underline{20.67} & 12.21 & \underline{26.67} & \underline{24.21} & \underline{29.53} \\
    Mean (Zero-Shot) & 26.88 & 23.40 & \underline{9.92} & 39.01 & 42.25 & 46.48 \\
    \bottomrule
    \end{tabularx}
\end{table}

Sub-millimeter values when using GT 2D confirm geometric consistency within floating-point precision. However, detector-based reconstructions remain in the 20–40~mm range because alignment corrects translation and rotation but not inter-view noise.

\begin{table*}[t]
    \centering
    \scriptsize
    \setlength{\tabcolsep}{4pt}
    \caption{\textbf{Simple Baselines for Monocular 3D Spine Pose Lifting.} Evaluation of \citeauthor{martinez2017simple}’s lifting model trained on spine-only (15 joints) and full-body (37 joints) keypoint sets. Reported metric: Procrustes-aligned MPJPE (P-MPJPE, mm) per action. Training on full-body joints improves spine localization accuracy. Evaluation is on 15 spine joints only.}
    \label{tab:baselines-spine3d-monocular-detailed}
    \begin{tabularx}{\linewidth}{lXccccccccccccccccccc}
    \toprule
    \textbf{Train Set} &
    \textbf{2D} &
    \textbf{Direct} &
    \textbf{Discuss} &
    \textbf{Eating} &
    \textbf{Greet} &
    \textbf{Phone} &
    \textbf{Photo} &
    \textbf{Pose} &
    \textbf{Purchase} &
    \textbf{Sit} &
    \textbf{SitD} &
    \textbf{Smoke} &
    \textbf{Wait} &
    \textbf{WalkD} &
    \textbf{WalkT} &
    \textbf{Walk} &
    \textbf{Avg} \\
    \midrule
    Spine Only & Det. & 16.58 & 17.92 & 17.79 & 18.32 & 21.81 & 19.74 & 15.69 & 17.70 & 19.28 & 22.32 & 18.43 & 18.82 & 18.91 & 16.24 & 15.96 & 18.58 \\
    Full-Body & Det. & \underline{15.55} & \underline{16.40} & \underline{14.45} & \underline{15.98} & \underline{19.84} & \underline{17.66} & \underline{14.25} & \underline{15.68} & \underline{15.97} & \underline{19.23} & \underline{16.38} & \underline{17.17} & \underline{17.41} & \underline{12.24} & \underline{12.12} & \underline{16.28} \\
    \midrule
    Spine Only & GT & 15.66 & 17.01 & 16.05 & 17.43 & 20.64 & 18.34 & 14.87 & 16.40 & 17.91 & 22.10 & 16.92 & 17.78 & 17.96 & 15.39 & 15.28 & 17.52 \\ 
    Full-Body & GT & \underline{11.56} & \underline{13.05} & \underline{11.36} & \underline{13.61} & \underline{18.06} & \underline{14.61} & \underline{10.60} & \underline{12.39} & \underline{13.64} & \underline{17.36} & \underline{13.25} & \underline{14.47} & \underline{14.51} & \underline{9.15} & \underline{9.98} & \underline{13.48} \\
    \bottomrule
    \end{tabularx}
\end{table*}

\begin{table}[t]
    \centering
    \scriptsize
    \setlength{\tabcolsep}{4pt}
    \caption{\textbf{Monocular 3D Spine Pose Lifting (Component-Wise).}
    Same experimental setup as Table~\ref{tab:baselines-spine3d-monocular-detailed}, but results are aggregated across actions and broken down by spine components.
    Columns report P-MPJPE (mm) for $S_C$ (cervical), $S_T$ (thoracic), $S_L$ (lumbar), and $S$ (full spine).
    Rows compare training on spine-only (15 joints) vs. full-body (37 joints); 2D inputs are either detected (Det.) or ground truth (GT).}
    \label{tab:baselines-spine3d-monocular}
    \begin{tabularx}{\linewidth}{lXYYYYYYYY}
    \toprule
    & & \multicolumn{4}{c}{\textbf{P-MPJPE}} &
    \multicolumn{4}{c}{\textbf{MPJPE}} \\
    \cmidrule(lr){3-6} \cmidrule(lr){7-10}
    \textbf{Train Set} &
    \textbf{2D} &
    \textbf{${\mathrm{S_C}}$} &
    \textbf{${\mathrm{S_T}}$} &
    \textbf{${\mathrm{S_L}}$} &
    \textbf{${\mathrm{S}}$} &
    \textbf{${\mathrm{S_C}}$} &
    \textbf{${\mathrm{S_T}}$} &
    \textbf{${\mathrm{S_L}}$} &
    \textbf{${\mathrm{S}}$} \\
    \cmidrule(lr){1-2} \cmidrule(lr){3-6} \cmidrule(lr){7-10}
    Spine Only & Det. & 12.98 & 14.29 & 3.32 & 18.58 &  \underline{96.88} & 75.96 & 18.83 & 67.70 \\
    Full-Body & Det. & \underline{12.81} & \underline{10.12} & \underline{3.51} & \underline{16.28} & 100.50 & \underline{72.61} & \underline{18.16} & \underline{67.39} \\
    \cmidrule(lr){1-2} \cmidrule(lr){3-6} \cmidrule(lr){7-10}
    Spine Only & GT &  12.03 & 13.45 & 3.32 & 17.52 & 77.93 & 58.70 & 15.54 & 53.60 \\
    Full-Body & GT & \underline{10.94} & \underline{7.98} & \underline{2.03} & \underline{13.48} & \underline{39.24} & \underline{26.97} & \underline{7.76} & \underline{25.94} \\
    \bottomrule
    \end{tabularx}
\end{table}

This experiment quantifies the geometric upper bound achievable conditioned on 2D detection accuracy alone.

\subsection{Monocular 3D Lifting}

\paragraph{Definition}
Learn
\[
f_{\text{lift}}:\R^{B \times T \times K \times 2} \rightarrow \R^{B \times T \times (K-1) \times 3},
\]
mapping 2D keypoints to root-centered 3D. Let $\mathcal{R}$ be the root index set (here $\mathcal{R}{=}\{0\}$). For frame $t$,
$\bar{\mathbf{y}}_{b,t} {=} \frac{1}{|\mathcal{R}|}\sum_{r\in\mathcal{R}} \mathbf{Y}_{b,t,r,:}$ and $\widetilde{\mathbf{Y}}_{b,t,j,:} {=} \mathbf{Y}_{b,t,j,:} {-} \bar{\mathbf{y}}_{b,t}$; the network minimizes $\|\widehat{\widetilde{\mathbf{Y}}} {-} \widetilde{\mathbf{Y}}\|_2^2$.

\paragraph{Baselines}
We adopt SimpleBaseline3D~\cite{martinez2017simple} as the reference architecture due to its simplicity and interpretability. It encodes per-joint features through linear layers with residual connections, operating on either spine-only or full-body keypoints.


\paragraph{Training} 
3D targets are root-centered and standardized per joint. The root joint is excluded during training and reinserted at inference. Frames are sampled at 1\,Hz to preserve motion diversity while maintaining manageable sequence length. Optimization uses AdamW~\cite{loshchilov2019adamw} with step scheduling. As in \cite{martinez2017simple}, a standard MSE loss is used for training.


\paragraph{Evaluation}
We report root-relative Mean Per-Joint Position Error (MPJPE) and Procrustes-aligned MPJPE (P-MPJPE) on decoded 3D predictions.
For P-MPJPE, each predicted frame is first aligned to the ground truth using the Procrustes similarity alignment $(s,\mathbf{R},\mathbf{t})$. All metrics are reported in millimeters. \cref{tab:baselines-spine3d-monocular-detailed} summarizes the P-MPJPE of spine keypoints across all dataset actions, and \cref{tab:baselines-spine3d-monocular} compares both P-MPJPE and root-relative MPJPE with further breakdown of three key spinal regions. Exact keypoints evaluated in each region are in Supplementary Sec. B.






\subsection{Ablation Studies}
\label{sec:ablations}

We report here controlled studies isolating the effect of data mixing, sampling, and training choices.

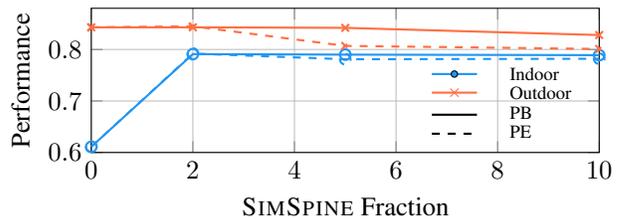
\begin{figure}[hb]
\centering
\begin{subfigure}{\columnwidth}
\begin{tikzpicture}
\begin{axis}[
    xlabel={\dataset Fraction}, ylabel={Performance},
    width=\linewidth, height=3.5cm,
    xmin=0, xmax=10, ymin=0.6, ymax=0.88,
    grid=both,
]
\addplot+[
    mark=o,
    color=easyblue,
    thick
] coordinates {
    (0,0.611)
    (2,0.791)
    (5,0.790)
    (10,0.789)
};

\addplot+[
    mark=o,
    color=easyblue,
    thick,
    dashed
] coordinates {
    (0,0.611)
    (2,0.792)
    (5,0.781)
    (10,0.782)
};

\addplot+[
    mark=x,
    color=hardorange,
    thick
] coordinates {
    (0,0.843)
    (2,0.843)
    (5,0.842)
    (10,0.828)
};

\addplot+[
    mark=x,
    color=hardorange,
    thick,
    dashed
] coordinates {
    (0,0.843)
    (2,0.845)
    (5,0.807)
    (10,0.801)
};

\node[anchor=south west, font=\scriptsize] at (axis cs:6.5,0.6) {
    \begin{tabular}{@{}ll@{}}
    \raisebox{1pt}{\tikz{\draw[thick, easyblue] (0,0)--(6mm,0); \draw[fill=easyblue] (3mm,0) circle (1.2pt);}} & Indoor \\
    \raisebox{1pt}{\tikz{\draw[thick, hardorange] (0,0)--(6mm,0); \draw[mark=x, mark size=2.2pt, hardorange, only marks] plot coordinates {(3mm,0)};}} & Outdoor \\
    \raisebox{1pt}{\tikz{\draw[thick] (0,0)--(6mm,0);}} & PB \\
    \raisebox{1pt}{\tikz{\draw[thick, dashed] (0,0)--(6mm,0);}} & PE
    \end{tabular}
};
\end{axis}
\end{tikzpicture}
\end{subfigure}
\caption{\textbf{Ablation Study: Mixup Composition.} 
We examine how the fraction of \dataset used in training influences indoor (AUC) and outdoor (AP) performance. Increasing the \dataset fraction improves indoor performance up to 10\%, while outdoor gains saturate by 2--5\%. 
Per-batch (PB) mixup maintains the best balance between indoor and outdoor metrics, whereas per-epoch (PE) alternation favors one domain at the expense of the other. Sampling only 2\% of \dataset achieves near-saturated results on both datasets, indicating diminishing returns from larger fractions.}
\label{fig:ablation-studies}
\end{figure}

\paragraph{Mixup Composition and Dataset Fraction}
Figure~\ref{fig:ablation-studies} analyzes how the fraction of \dataset contributes to 2D fine-tuning performance under different data mixing strategies. This ablation is motivated by the need for stable and accurate 2D keypoint detectors, which form the foundation for reliable 3D triangulation and lifting. Existing off-the-shelf 2D detectors trained on manually annotated spine datasets often exhibit label noise and jitter due to inconsistent supervision, making it essential to study how synthetic and real data can be best combined. Increasing the proportion of \dataset from 0\% to 10\% steadily improves indoor AUC (from 0.61 to 0.79), with minimal degradation on outdoor AP. Beyond 2--5\%, however, outdoor AP plateaus around 0.84, suggesting that only a small subset of high-quality indoor samples is sufficient to enhance generalization. Comparing mixing strategies, per-batch (PB) mixup consistently outperforms per-epoch (PE) alternation, yielding the most stable trade-off across domains. PB works better because the model and optimizer jointly observe samples from both datasets in each iteration, allowing AdamW to maintain smoother gradient statistics and consistent momentum estimates across domains. These results justify using 2\% of \dataset—about 31k indoor images, roughly matching the 33k outdoor samples in SpineTrack—with per-batch mixing as the default configuration, achieving balanced exposure across domains and efficient use of synthetic data while preserving real-world performance.

\subsection{Discussion}
\label{sec:discussion}

Across our three experiments—2D detection, multiview triangulation, and monocular lifting—we establish reference baselines for spine‑aware pose estimation. In 2D, fine‑tuning across SpineTrack and \dataset consistently improves \emph{spine} metrics (indoor AUC from 0.61 to 0.80; SpineTrack AP$^{\mathrm{S}}$ from 0.91 to 0.93), with a small trade‑off on body AP$^{\mathrm{B}}$ relative to the strongest SpinePose pretrained model. The multiview triangulation baseline attains \textbf{31.8\,mm} MPJPE and \textbf{29.5\,mm} P‑MPJPE, and our oracle using GT 2D reaches sub‑millimeter P‑MPJPE, confirming geometric consistency. For monocular lifting, the full‑body variant outperforms the spine‑only variant (detected 2D: 18.6\,mm $\rightarrow$ \underline{16.3}\,mm P‑MPJPE; GT 2D: 17.5\,mm $\rightarrow$ \underline{13.5}\,mm), indicating that global context aids vertebral localization. Together, these results provide a practical baseline suite that links strong 2D cues to geometrically faithful 3D reconstructions and highlights where temporal or biomechanical priors add the most value.

\paragraph{Limitations}
Our simulation framework is a kinematics-only, simulation-derived resource and carries several limitations. Anatomically, we articulate five intervertebral lumbar joints and a single 3‑DOF cervicothoracic joint, while treating the remaining thoracic and cervical segments as rigid. This choice, made for numerical stability and identifiability from RGB inputs, neglects rib‑cage coupling and soft‑tissue effects that materially constrain and distribute thoracic motion~\cite{liebsch2018thoracic}. Intervertebral translations are not modeled; although small in healthy spines, nonzero translations have been measured in vivo with stereoradiography~\cite{pearcy1985stereo,pearcy1984three}. Pelvis–lumbar coupling is also simplified, which can underrepresent lumbopelvic rhythm during trunk motion~\cite{tafazzol2014lumbopelvic}. Subjects are implicitly healthy and scaled by height/mass; age-, sex-, and pathology‑dependent variation in sagittal alignment is not modeled, so the dataset encodes nominal healthy‑motion priors rather than the diversity seen clinically~\cite{ludwig2023reference,zappala2021relationship}. Because all motions are sourced from Human3.6M, the visual domain is limited to indoor captures with fixed multi‑view cameras and a restricted action set~\cite{ionescu2014h36m}. This may limit generalization to fully unconstrained, in‑the‑wild scenarios. Our OpenSim step solves inverse kinematics only; muscle actuation, ground‑reaction forces, and load equilibria are not enforced, so trajectories are geometrically plausible but not physically validated~\cite{hicks2015bestpractices}. Extending to dynamics‑consistent optimal control would allow force‑consistent motion generation~\cite{dembia2020opensim}. Our empirical validation (\cref{sec:validation}) target geometric plausibility (curvature envelopes, ROM profiles) rather than absolute accuracy against in vivo ground truth such as dual‑fluoroscopy/biplane tracking or standing biplanar reconstructions, which provide the most precise in vivo vertebral kinematics and alignment~\cite{anderst2011validation,melhem2016eos}. Accordingly, \framework framework should be regarded as a scalable proxy for method development and benchmarking rather than clinical measurement, while \dataset serves best as a large-scale pretraining resource for spine pose estimation models later fine-tuned on smaller, biomechanically validated datasets.

\paragraph{Future work} Angular annotations in \dataset, not evaluated in this work because of additional design choices (representation, normalization, interpretation) orthogonal to this initial benchmark, should be benchmarked. In addition, we see three priorities: (i) expand anatomical fidelity by adding rib‑cage articulation and small intervertebral translations, (ii) couple IK with dynamics (inverse dynamics or optimal control) to ensure force‑consistent motion~\cite{dembia2020opensim}, and (iii) broaden the visual domain with in‑the‑wild sequences and pathology‑specific cohorts. Longer term, combining upright clinical imaging (EOS/fluoroscopy) with RGB could anchor subject‑specific spine priors and reduce ambiguity in vertebra‑level motion.

\section{Conclusion}
\label{sec:conclusion}

We introduced \framework, a pipeline that fuses calibrated RGB, subject‑scaled musculoskeletal models, and virtual vertebral markers to produce anatomically constrained 3D spine motion from standard multi‑view footage. The resulting \dataset provides 15 keypoints driving spine with per‑segment rotations over 2.14M frames and a set of pretrained baselines spanning 2D detection, multiview triangulation, and monocular lifting. The evidence—near‑oracle P‑MPJPE under GT 2D, consistent thoracolumbar curvature envelopes, and action‑sensitive ROM profiles—shows that simulation‑driven annotation can extend existing datasets with biomechanically meaningful structure. While simplified and domain‑limited, the framework offers a practical bridge between computer vision and musculoskeletal modeling, enabling models that reason about posture and vertebral kinematics, not only joint geometry.

\paragraph{Availability and licensing}
Code, models, and \dataset annotations for the spine markers and kinematic parameters will be released for research use only. Due to licensing, full‑body keypoints are reproducible by running our pipeline on the licensed \human data.

\section*{Acknowledgement}

This work was co-funded by the European Union’s Horizon Europe research and innovation programme under Grant Agreement No 101135724 (LUMINOUS) and Grant Agreement No 101092889 (SHARESPACE).

{
    \small
    \bibliographystyle{ieeenat_fullname}
    \bibliography{main}
}


\end{document}